\documentclass[10pt,twocolumn,letterpaper]{article}
\usepackage{cvpr}
\usepackage{times}
\usepackage{epsfig}
\usepackage{graphicx}
\usepackage{amsmath}
\usepackage{amssymb}

\usepackage[utf8]{inputenc} 
\usepackage[T1]{fontenc}    
\usepackage{url}            
\usepackage{booktabs}       
\usepackage{amsfonts}       
\usepackage{nicefrac}       
\usepackage{microtype}      
\usepackage[numbers]{natbib}

\usepackage[bookmarks=false]{hyperref}
\usepackage{xcolor}

\hypersetup{
  colorlinks,
  breaklinks=true,
  anchorcolor={blue!50!black},
  linkcolor={blue!50!black},
  citecolor={blue!50!black},
  urlcolor={blue}
}

\usepackage{color}
\usepackage{xspace}
\usepackage{sidecap}
\usepackage{floatrow}
\usepackage{subcaption}
\usepackage{epsfig}
\usepackage{multirow}
\usepackage{epsfig}
\usepackage{comment}
\usepackage{amsthm}
\usepackage{gensymb}
\sidecaptionvpos{figure}{c}

\usepackage{amsmath,amsfonts,bm}

\def\eqref#1{equation~\ref{#1}}

\def\1{\bm{1}}

\DeclareMathAlphabet{\mathsfit}{\encodingdefault}{\sfdefault}{m}{sl}
\SetMathAlphabet{\mathsfit}{bold}{\encodingdefault}{\sfdefault}{bx}{n}

\DeclareMathOperator{\Tr}{Tr}

\DeclareMathOperator{\FID}{FID} 

\newcommand{\celebahq}{\textsc{celeba-hq}}
\newcommand{\lsun}{\textsc{lsun-bedroom}}
\newcommand{\cifar}{\textsc{cifar10}}
\newcommand{\imagenet}{\textsc{imagenet}}

\cvprfinalcopy 

\def\cvprPaperID{5940} 

\ifcvprfinal\fi
\begin{document}

\title{Self-Supervised GANs via Auxiliary Rotation Loss}

\author{
Ting Chen\thanks{Work done at Google.}\\
University of California, Los Angeles\\
{\tt\small tingchen@cs.ucla.edu}
\and
Xiaohua Zhai\\
Google Brain\\
{\tt\small xzhai@google.com}
\and
Marvin Ritter\\
Google Brain\\
{\tt\small marvinritter@google.com}
\and
Mario Lucic\\
Google Brain\\
{\tt\small lucic@google.com}
\and
Neil Houlsby\\
Google Brain\\
{\tt\small neilhoulsby@google.com}
}

\maketitle

\begin{abstract}
  Conditional GANs are at the forefront of natural image synthesis. The main drawback of such models is the necessity for labeled data. In this work we exploit two popular unsupervised learning techniques, adversarial training and self-supervision, and take a step towards bridging the gap between conditional and unconditional GANs. In particular, we allow the networks to collaborate on the task of representation learning, while being adversarial with respect to the classic GAN game. The role of self-supervision is to encourage the discriminator to learn meaningful feature representations which are not forgotten during training. We test empirically both the quality of the learned image representations, and the quality of the synthesized images. Under the same conditions, the self-supervised GAN attains a similar performance to state-of-the-art conditional counterparts. Finally, we show that this approach to fully unsupervised learning can be scaled to attain an FID of 23.4 on unconditional \imagenet{} generation.\footnote{Code at \url{https://github.com/google/compare_gan}.}
\end{abstract}

\section{Introduction}
\begin{figure*}[t!]
\includegraphics[width=0.75\textwidth]{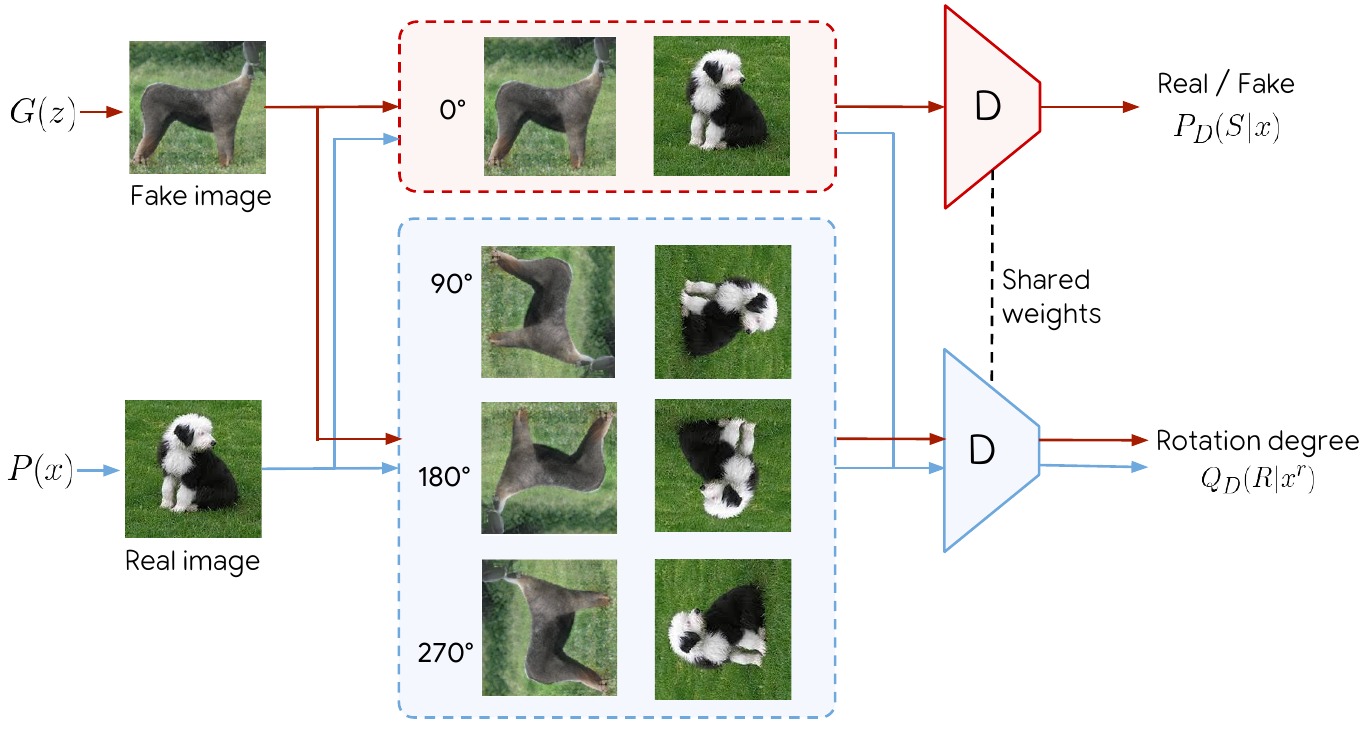}
\caption{
Discriminator with rotation-based self-supervision.
The discriminator, $D$, performs two tasks: true vs. fake binary classification, and rotation degree classification. Both the fake and real images are rotated by 0, 90, 180, and 270 degrees. The colored arrows indicate that only the upright images are considered for true vs. fake classification loss task. For the rotation loss, all images are classified by the discriminator according to their rotation degree.
}
\label{fig:rotation_acgan}
\end{figure*}

Generative Adversarial Networks (GANs) are a class of unsupervised generative models~\citep{goodfellow2014generative}.
GANs involve training a \emph{generator} and \emph{discriminator} model in an adversarial game, such that the generator learns to produce samples from a desired data distribution.
Training GANs is challenging because it involves searching for a Nash equilibrium of a non-convex game in a high-dimensional parameter space. In practice, GANs are typically trained using alternating stochastic gradient descent which is often unstable and lacks theoretical guarantees~\cite{salimans2016improved}. Consequently, training may exhibit instability, divergence, cyclic behavior, or mode collapse~\citep{mescheder2018training}. As a result, many techniques to stabilize GAN training have been proposed~\citep{mao2016least,gulrajani2017improved,miyato2018spectral,abn,radford2016,zhang2018self,karras2017progressive}. A major contributor to training instability is the fact that the generator and discriminator learn in a non-stationary environment. In particular, the discriminator is a classifier for which the distribution of one class (the fake samples) shifts as the generator changes during training. In non-stationary online environments, neural networks forget previous tasks~\cite{kirkpatrick2017overcoming,mccloskey1989catastrophic,french1999catastrophic}. If the discriminator forgets previous classification boundaries, training may become unstable or cyclic. This issue is usually addressed either by reusing old samples or by applying continual learning techniques~\citep{thanh2018catastrophic,anonymous2019generative,seff2017continual,shrivastava2017learning,kim2018memorization,grnarova2017online}.
These issues become more prominent in the context of complex data sets. A key technique in these settings is \emph{conditioning}~\citep{zhang2018self,odena2017,miyato2018cgans,brock2018large} whereby
both the generator and discriminator have access to labeled data. Arguably, augmenting the discriminator with supervised information encourages it to learn more stable representations which opposes catastrophic forgetting. Furthermore, learning the conditional model for each class is easier than learning the joint distribution. The main drawback in this setting is the necessity for labeled data. Even when labeled data is available, it is usually sparse and covers only a limited amount of high level abstractions.

Motivated by the aforementioned challenges, our goal is to show that one can recover the benefits of conditioning, \emph{without requiring labeled data}. To ensure that the representations learned by the discriminator are more stable and useful, we add an auxiliary, self-supervised loss to the discriminator. This leads to more stable training because the dependence of the discriminator's representations on the quality of the generator's output is reduced. We introduce a novel model -- the \emph{self-supervised} GAN -- in which the generator and discriminator collaborate on the task of representation learning, and compete on the generative task.

\noindent\textbf{Our contributions}\quad
We present an unsupervised generative model that combines adversarial training with self-supervised learning.
Our model recovers the benefits of conditional GANs, but requires no labeled data.
In particular, \emph{under the same training conditions}, the self-supervised GAN closes the gap in natural image synthesis between unconditional and conditional models. Within this setting the quality of discriminator's representations is greatly increased which might be of separate interest in the context of transfer learning. A large-scale implementation of the model leads to promising results on \emph{unconditional} \imagenet{} generation, a task considered daunting by the community. We believe that this work is an important step in the direction of high quality, fully unsupervised, natural image synthesis.

\section{A Key Issue: Discriminator Forgetting}

The original value function for GAN training is~\citep{goodfellow2014generative}:
\begin{equation}
\label{eq:gan_obj_classic}
\begin{aligned}
V(G, D) =& \mathbb{E}_{\bm x\sim P_{\mathrm{data}}(\bm x)}[\log P_D(S=1\mid\bm x)] \\
&+ \mathbb{E}_{\bm x\sim P_{G}(\bm x)}[\log (1- P_D(S=0\mid\bm x))]
\end{aligned}
\end{equation}
where $P_{\mathrm{data}}$ is the true data distribution, and $P_G$ is the distribution induced by transforming a simple distribution $\bm z\sim P(\bm z)$ using the deterministic mapping given by the generator, $\bm x = G(\bm z)$, and $P_D$ is the discriminator's Bernoulli distribution over the labels (true or fake). In the original minimax setting the generator maximizes Equation~\ref{eq:gan_obj_classic} with respect to it's parameters, while the discriminator minimizes it. Training is typically performed via alternating stochastic gradient descent. Therefore, at iteration $t$ during training, the discriminator classifies samples as coming from $P_\mathrm{data}$ or  $P_G^{(t)}$. As the parameters of G change, the distribution $P_G^{(t)}$ changes, which implies a non-stationary online learning problem for the discriminator.

This challenge has received a great deal of attention and explicit temporal dependencies have been proposed to improve training in this setting~\citep{salimans2016improved,anonymous2019generative,shrivastava2017learning,grnarova2017online}. Furthermore, in online learning of non-convex functions, neural networks have been shown to forget previous tasks~\citep{kirkpatrick2017overcoming,mccloskey1989catastrophic,french1999catastrophic}. In the context of GANs, learning varying levels of detail, structure, and texture, can be considered different tasks. For example, if the generator first learns the global structure, the discriminator will naturally try to build a representation which allows it to efficiently penalize the generator based only on the differences in global structure, or the lack of local structure. As such, one source of instability in training is that the discriminator is not incentivised to maintain a useful data representation as long as the current representation is useful to discriminate between the classes.

Further evidence can be gathered by considering the generator and discriminator at convergence. Indeed, \citet{goodfellow2014generative} show that the optimal discriminator estimates the likelihood ratio between the generated and real data distributions. Therefore, given a perfect generator, where $P_G = P_\mathrm{data}$, the optimal discriminator simply outputs $0.5$, which is a constant and doesn't depend on the input. Hence, this discriminator would have no requirement to retain meaningful representations. Furthermore, if regularization is applied, the discriminator might ignore all but the minor features which distinguish real and fake data.

\begin{figure}[t]
  \centering
  \includegraphics[width=0.8\textwidth]{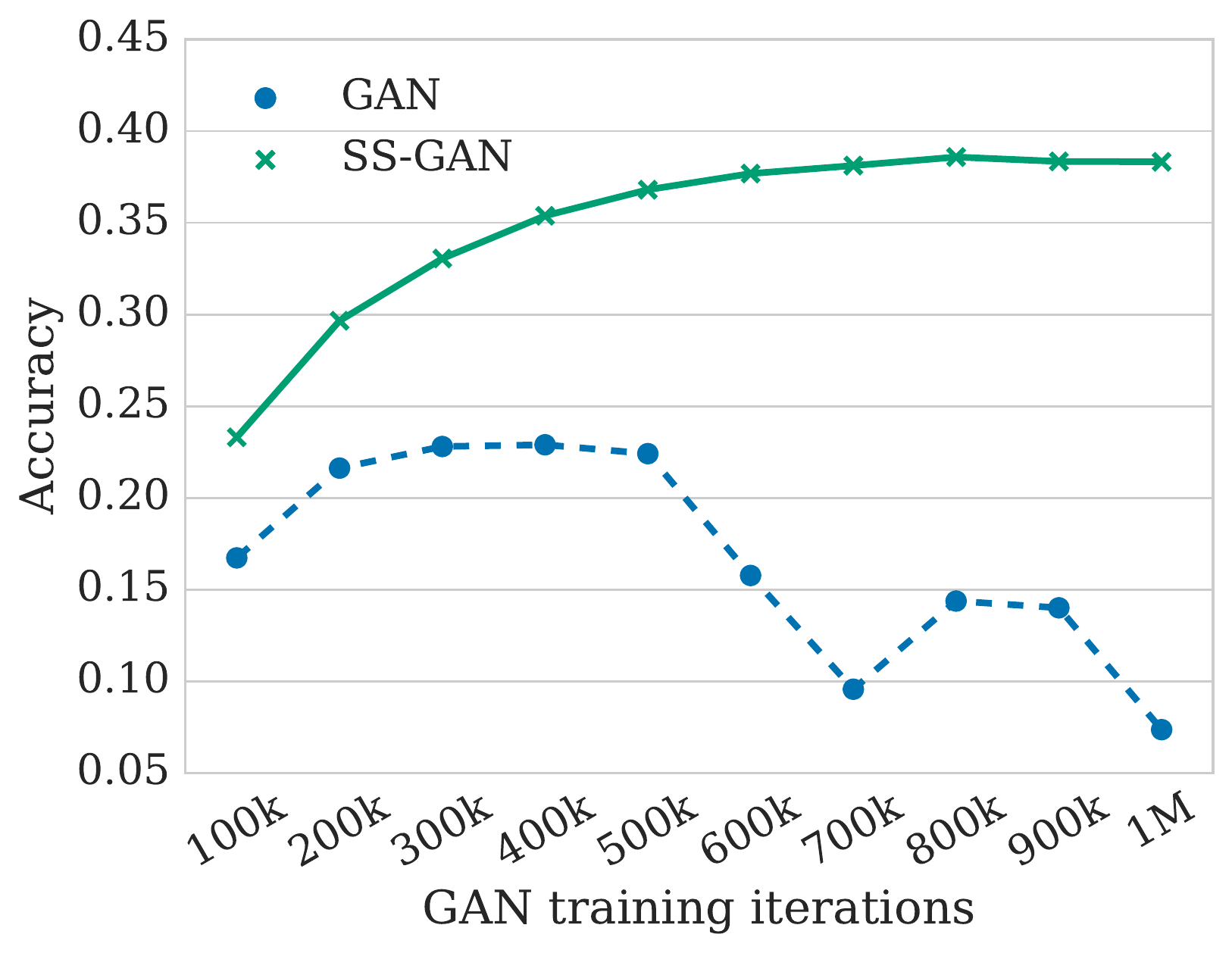}
  \caption{
  \label{fig:gan_forget}
  Performance of a linear classification model, trained on \imagenet{} on representations extracted from the final layer of the discriminator.
  Uncond-GAN denotes an unconditional GAN. SS-GAN denotes the same model when self-supervision is added. For the Uncond-GAN, the representation gathers information about the class of the image and the accuracy increases.
  However, after $500$k iterations, the representations lose information about the classes and performance decreases.
  SS-GAN alleviates this problem. More details are presented in Section~\ref{sec:experiments}.}
\end{figure}

\begin{figure}[t]
\begin{center}
\epsfig{file=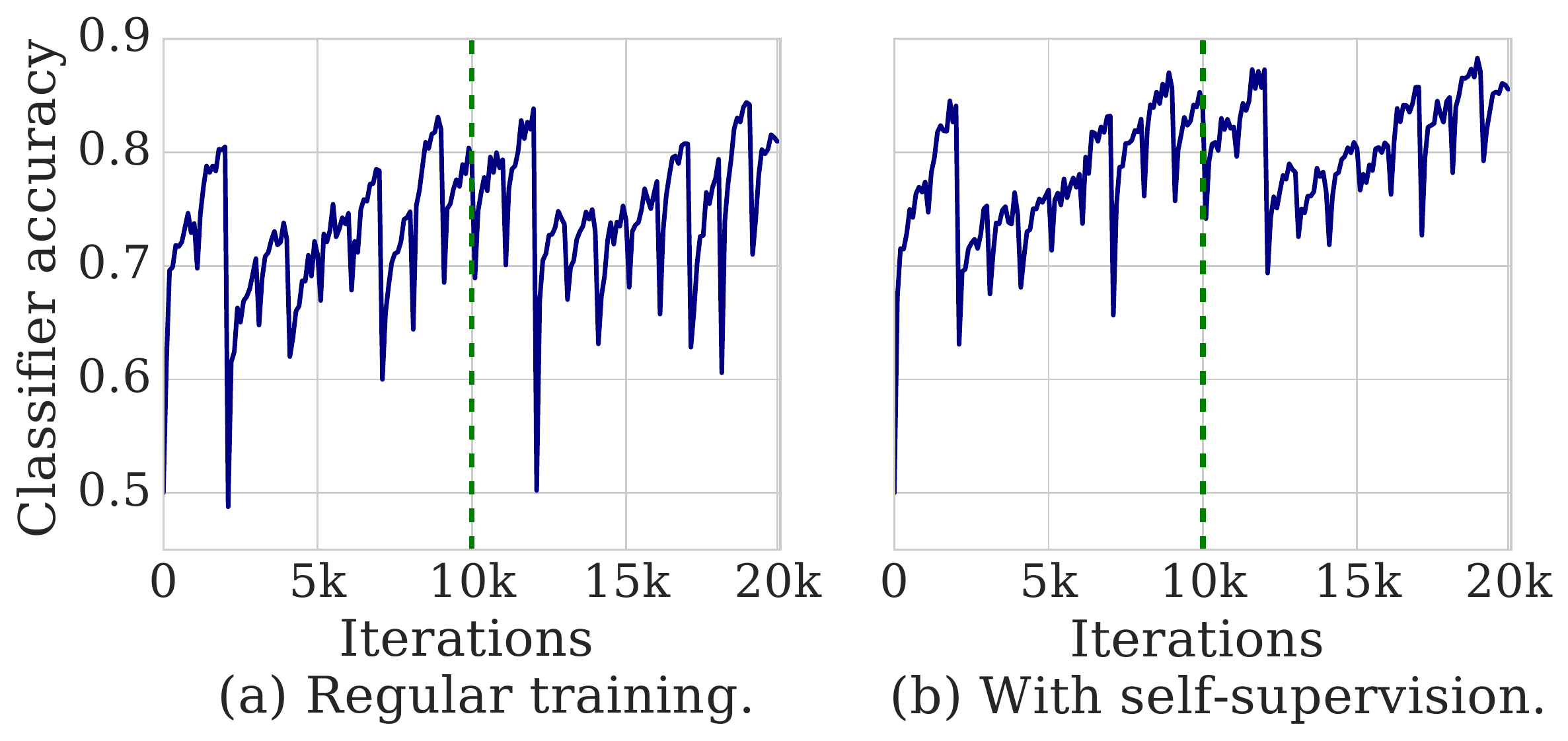,height=4cm}
\end{center}
\caption{\label{fig:non-stationary}
Image classification accuracy when the underlying class distribution shifts every $1$k iterations.
The vertical dashed line indicates the end of an entire cycle through the tasks, and return to the original classification task at $t=0$.
\emph{Left}: vanilla classifier.
\emph{Right}: classifier with an additional self-supervised loss.
This example demonstrates that a classifier may fail to learn generalizable representations in a non-stationary environment, but self-supervision helps mitigate this problem.}
\end{figure}

We demonstrate the impact of discriminator forgetting in two settings.
(1) A simple scenario shown in Figure~\ref{fig:non-stationary}(a), and,
(2) during the training of a GAN shown in Figure~\ref{fig:gan_forget}.
In the first case a classifier is trained sequentially on 1-vs.-all classification tasks on each of the ten classes in \textsc{cifar10}. It is trained for $1$k iterations on each task before switching to the next. At $10$k iterations the training cycle repeats from the first task. Figure~\ref{fig:non-stationary}(a) shows substantial forgetting, despite the tasks being similar.
Each time the task switches, the classifier accuracy drops substantially.
After $10$k iterations, the cycle of tasks repeats, and the accuracy is the same as the first cycle.
No  useful information is carried across tasks.
This demonstrates that the model does not retain generalizable representations in this non-stationary environment. In the second setting shown in Figure~\ref{fig:gan_forget} we observe a similar effect during GAN training. Every $100$k iterations, the discriminator representations are evaluated on \imagenet{} classification; the full protocol is described in Section~\ref{sec:self_sup_eval}.
During training, classification of the unconditional GAN increases, then decreases, indicating that information about the classes is acquired and later forgotten.
This forgetting correlates with training instability.
Adding self-supervision, as detailed in the following section, prevents this forgetting of the classes in the discriminator representations.

\section{The Self-Supervised GAN}
Motivated by the main challenge of discriminator forgetting, we aim to imbue the discriminator with a mechanism which allows learning useful representations, independently of the quality of the current generator. To this end, we exploit recent advancements in self-supervised approaches for representation learning. The main idea behind self-supervision is to train a model on a pretext task like predicting rotation angle or relative location of an image patch, and then extracting representations from the resulting networks~\citep{dosovitskiy2014,doersch2015unsupervised,zhang2016colorful}. We propose to add a self-supervised task to our discriminator.

In particular, we apply the state-of-the-art self-supervision method based on image rotation~\citep{gidaris2018unsupervised}.
In this method, the images are rotated, and the angle of rotation becomes the artificial label (cf. Figure~\ref{fig:rotation_acgan}).
The self-supervised task is then to predict the angle of rotation of an image. The effects of this additional loss on the image classification task is evident in Figure~\ref{fig:non-stationary}(b): When coupled with the self-supervised loss, the network learns representations that transfer across tasks and the performance continually improves.
On the second cycle through the tasks, from $10$k iterations onward, performance is improved.
Intuitively, this loss encourages the classifier to learn useful image representations to detect the rotation angles, which transfers to the image classification task.

We augment the discriminator with a rotation-based loss which results in the following loss functions:
\begin{align*}
L_G &= -V(G, D) -\alpha \mathbb{E}_{\bm x \sim P_G}\mathbb{E}_{r \sim \mathcal{R}}\left[\log Q_{D}(R=r\mid{\bm x}^r)\right],\\
L_D &= V(G, D) -\beta \mathbb{E}_{\bm x \sim P_{\textrm{data}}}\mathbb{E}_{r \sim \mathcal{R}}\left[\log Q_{D}(R=r\mid{\bm x}^r)\right],
\end{align*}
where $V(G, D)$ is the value function from Equation~\ref{eq:gan_obj_classic},
$r\in\mathcal{R}$ is a rotation selected from a set of possible rotations.
In this work we use $\mathcal{R} = \{ 0^{\degree}, 90^{\degree}, 180^{\degree}, 270^{\degree} \}$
as in~\citet{gidaris2018unsupervised}. Image $\bm x$ rotated by $r$ degrees is denoted as $\bm x^r$, and $Q(R \mid \bm x^r)$ is the discriminator's predictive distribution over the angles of rotation of the sample.

\paragraph{Collaborative Adversarial Training}
In our model, the generator and discriminator are adversarial with respect to the true vs. fake prediction loss, $V(G,D)$, however, they are \emph{collaborative} with respect to the rotation task.
First, consider the value function of the generator which biases the generation towards images, that when rotated, the discriminator can detect their rotation angle. Note that the generator is not conditional but only generates ``upright'' images which are subsequently rotated and fed to the discriminator. On the other hand, the discriminator is trained to detect rotation angles based \emph{only on the true data}. In other words, the parameters of the discriminator get updated only based on the rotation loss on the true data. This prevents the undesirable collaborative solution whereby the generator generates images whose subsequent rotation is easy to detect. As a result, the generator is encouraged to generate images that are rotation-detectable because they share features with real images that are used for rotation classification.

In practice, we use a single discriminator network with two heads to compute $P_D$ and $Q_{D}$. Figure~\ref{fig:rotation_acgan} depicts the training pipeline. We rotate the real and generated images in four major rotations. The goal of the discriminator on \textit{non-rotated} images is to predict whether the input is true or fake. On rotated \emph{real} images, its goal is to detect the rotation angle. The goal of the generator is to generate images matching the observed data, whose representation in the feature space of the discriminator allows detecting rotations. With $\alpha>0$ convergence to the true data distribution $P_G=P_{\mathrm{data}}$ is not guaranteed. However, annealing $\alpha$ towards zero during training will restore the guarantees.

\section{Experiments}
\label{sec:experiments}

We demonstrate empirically that
(1) self-supervision improves the representation quality with respect to baseline GAN models, and that
(2) it leads to improved unconditional generation for complex datasets, matching the performance of conditional GANs, under equal training conditions.

\subsection{Experimental Settings}

\noindent\textbf{Datasets}\quad
We focus primarily on \imagenet{}, the largest and most diverse image dataset commonly used to evaluate GANs.
Until now, most GANs trained on \imagenet{} are conditional.
\imagenet{} contains $1.3$M training images and $50$k test images.
We resize the images to $128\times128\times3$ as done in~\citet{miyato2018cgans} and ~\citet{zhang2018self}.
We provide additional comparison on three smaller datasets, namely
\cifar{}, \celebahq{}, \lsun{}, for which unconditional GANs can be successfully trained.
The \lsun{} dataset~\citep{yu15lsun} contains $3$M images.
We partition these randomly into a test set containing approximately $30$k images and a train set containing the rest.
\celebahq{} contains $30$k images~\citep{karras2017progressive}.
We use the $128\times128\times3$ version obtained by running the code provided by the authors.\footnote{\url{https://github.com/tkarras/progressive\_growing\_of\_gans}.}
We use $3$k examples as the test set and the remaining examples as the training set.  \cifar{} contains $70$k images ($32\times32\times3$),
partitioned into $60$k training instances and $10$k test instances.

\begin{table}[t]
  \centering
  \caption{\label{tab:best_fid} Best FID attained across three random seeds. In this setting the proposed approach recovers most of the benefits of conditioning.\vspace{-5mm}}
  \begin{tabular}{lll}
\toprule
\textsc{Dataset} & \textsc{Method} &  \textsc{FID}   \\
\midrule
\multirow{4}{*}{\cifar} & Uncond-GAN &     19.73 \\
        & Cond-GAN &     \textbf{15.60} \\
        & SS-GAN &     17.11 \\
        & SS-GAN (sBN) &     {15.65} \\ \midrule
\multirow{4}{*}{\imagenet{}} & Uncond-GAN &     56.67 \\
        & Cond-GAN &     \textbf{42.07} \\
        & SS-GAN &     47.56 \\
        & SS-GAN (sBN) &     43.87 \\ \midrule
\multirow{3}{*}{\lsun{}} & Uncond-GAN &     16.02 \\
        & SS-GAN &     13.66 \\
        & SS-GAN (sBN) &     \textbf{13.30} \\ \midrule
\multirow{3}{*}{\celebahq} & Uncond-GAN &     \textbf{23.77} \\
        & SS-GAN &     26.11 \\
        & SS-GAN (sBN) &     24.36 \\
\bottomrule
\end{tabular}

\end{table}

{\renewcommand{\arraystretch}{1.2}
\begin{table*}[t]
\small\centering
\caption{\label{tab:gp_robustness_fid}
FID for unconditional GANs under different hyperparameter settings.
Mean and standard deviations are computed across three random seeds.
Adding the self-supervision loss reduces the sensitivity of GAN training to hyperparameters.}
\begin{tabular}{lcllcllll}
\toprule
                 &    &     &       &  & \multicolumn{2}{c}{\cifar{}} & \multicolumn{2}{c}{\imagenet{}} \\
\textsc{type} & $\lambda$ & $\beta_1$ & $\beta_2$ & \textsc{D Iters} &  \textsc{Uncond-GAN} &          \textsc{SS-GAN} &              \textsc{Uncond-GAN} &           \textsc{SS-GAN}     \\
\midrule
\multirow{6}{*}{\textsc{gradient penalty}} & \multirow{3}{*}{1} & \multirow{2}{*}{0.0} & \multirow{2}{*}{0.900} & 1 &  121.05 $\pm$ 31.44 &   \textbf{25.8 $\pm$ 0.71} &  183.36 $\pm$ 77.21 &   \textbf{80.67 $\pm$ 0.43} \\ 
                 &    &     &       & 2 &    28.11 $\pm$ 0.66 &  \textbf{26.98 $\pm$ 0.54} &    85.13 $\pm$ 2.88 &   \textbf{83.08 $\pm$ 0.38} \\
                 &    & 0.5 & 0.999 & 1 &    78.54 $\pm$ 6.23 &  \textbf{25.89 $\pm$ 0.33} &   104.73 $\pm$ 2.71 &   \textbf{91.63 $\pm$ 2.78} \\ \cline{2-9}
                 & \multirow{3}{*}{10} & \multirow{2}{*}{0.0} & \multirow{2}{*}{0.900} & 1 &  188.52 $\pm$ 64.54 &  \textbf{28.48 $\pm$ 0.68} &  227.04 $\pm$ 31.45 &    \textbf{85.38 $\pm$ 2.7} \\
                 &    &     &       & 2 &    29.11 $\pm$ 0.85 &  \textbf{27.74 $\pm$ 0.73} &  227.74 $\pm$ 16.82 &   \textbf{80.82 $\pm$ 0.64} \\
                 &    & 0.5 & 0.999 & 1 &  117.67 $\pm$ 17.46 &  \textbf{25.22 $\pm$ 0.38} &  242.71 $\pm$ 13.62 &  \textbf{144.35 $\pm$ 91.4} \\
\cline{1-9}
\multirow{3}{*}{\textsc{spectral norm}} & \multirow{3}{*}{0} & \multirow{2}{*}{0.0} & \multirow{2}{*}{0.900} & 1 &    87.86 $\pm$ 3.44 &   \textbf{19.65 $\pm$ 0.9} &    129.96 $\pm$ 6.6 &   \textbf{86.09 $\pm$ 7.66} \\ 
                 &    &     &       & 2 &    20.24 $\pm$ 0.62 &  \textbf{17.88 $\pm$ 0.64} &    80.05 $\pm$ 1.33 &   \textbf{70.64 $\pm$ 0.31} \\
                 &    & 0.5 & 0.999 & 1 &    86.87 $\pm$ 8.03 &  \textbf{18.23 $\pm$ 0.56} &  201.94 $\pm$ 27.28 &   \textbf{99.97 $\pm$ 2.75} \\
\bottomrule
\end{tabular}
\end{table*}}

\paragraph{Models\label{sec:exp_models}}
We compare the self-supervised GAN (SS-GAN) to two well-performing baseline models, namely (1) the unconditional GAN with spectral normalization proposed in~\citet{miyato2018spectral}, denoted Uncond-GAN,
and (2) the conditional GAN using the label-conditioning strategy and the Projection Conditional GAN (Cond-GAN)~\citep{miyato2018cgans}.
We chose the latter as it was shown to outperform the AC-GAN~\citep{odena2017}, and is adopted by the best performing conditional GANs~\citep{zhang2018self,mescheder2018training,brock2018large}.

We use ResNet architectures for the generator and discriminator as in~\citet{miyato2018spectral}.
For the conditional generator in Cond-GAN,
we apply label-conditional batch normalization.
In contrast, SS-GAN does not use conditional batch normalization.
However, to have a similar effect on the generator, we consider a variant of SS-GAN where we apply the self-modulated batch normalization which does not require labels~\citep{abn} and denote it SS-GAN (sBN).
We note that labels are available only for \cifar{} and \imagenet{}, so Cond-GAN is only applied on those data sets.

We use a batch size of 64 and to implement the rotation-loss we rotate 16 images in the batch in all four considered directions.
We do not add any new images into the batch to compute the rotation loss.
For the true vs. fake task we use the hinge loss from \citet{miyato2018spectral}.
We set $\beta=1$ or the the self-supervised loss.
For $\alpha$ we performed a small sweep $\alpha\in\{0.2, 0.5, 1\}$, and select $\alpha=0.2$ for all datasets (see the appendix for details).
For all other hyperparameters, we use the values in \citet{miyato2018spectral} and \citet{miyato2018cgans}.
We train \cifar{}, \lsun{} and \celebahq{} for $100$k steps on a single P100 GPU.
For \imagenet{} we train for $1$M steps.
For all datasets we use the Adam optimizer with learning rate $0.0002$.

\subsection{Comparison of Sample Quality \label{sec:fid_quantitative_compare}}
\noindent\textbf{Metrics}\quad
To evaluate generated samples from different methods quantitatively, we use the Frechet Inception Distance (FID)~\citep{heusel2017gans}.
In FID, the true data and generated samples are first embedded in a specific layer of a pre-trained Inception network.
Then, a multivariate Gaussian is fit to the data and the distance computed as
$\FID(x, g) = ||\mu_x -\mu_g||_2^2 + \Tr(\Sigma_x + \Sigma_g - 2(\Sigma_x\Sigma_g)^\frac12)$,
where
$\mu$ and $\Sigma$ denote the empirical mean and covariance and subscripts $x$ and $g$ denote the true and generated data respectively. FID is shown to be sensitive to both the addition of spurious modes and to mode dropping~\citep{sajjadi2018assessing,lucic2018}.
An alternative approximate measure of sample quality is Inceptions Score (IS)~\citet{salimans2016improved}.
Since it has some flaws~\citet{barratt2018note}, we use FID as the main metric in this work.

\begin{figure*}[ht]
 \centering
 \begin{subfigure}[b]{0.425\textwidth}
    \includegraphics[width=1\textwidth]{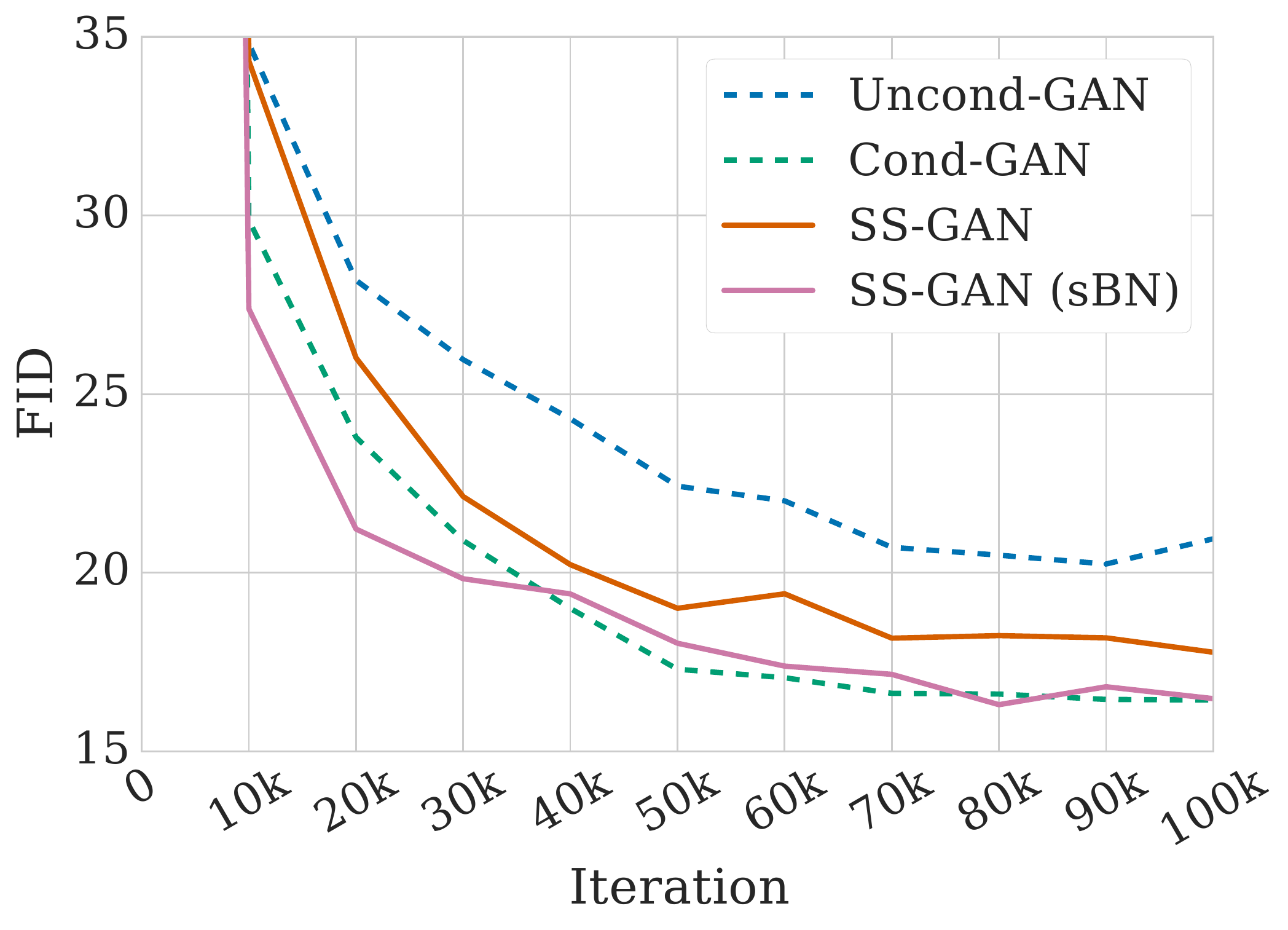}
    \caption{\cifar{}}
 \end{subfigure}
 \begin{subfigure}[b]{0.425\textwidth}
    \includegraphics[width=1\textwidth]{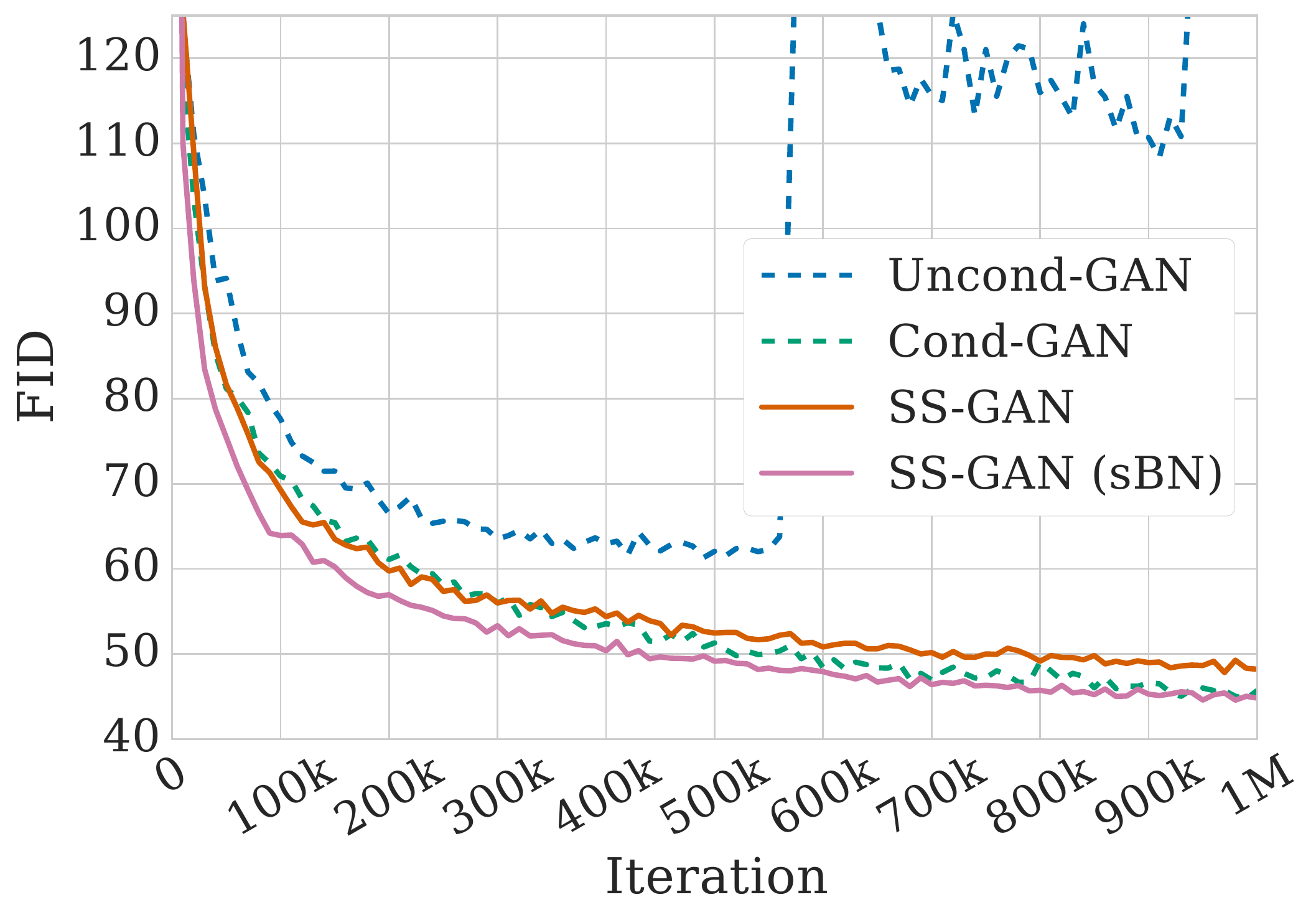}
    \caption{\imagenet{}}
 \end{subfigure}\hfill
 \caption{\label{fig:convergence_curves}
   FID learning curves on \cifar{} and \imagenet{}.
   The curves show the mean performance across three random seeds.
   The unconditional GAN (Uncond-GAN) attains significantly poorer performance than the conditional GAN (Cond-GAN).
   The unconditional GAN is unstable on \imagenet{} and the runs often diverge after $500$k training iterations.
   The addition of self-supervision (SS-GAN) stabilizes Uncond-GAN and boosts performance.
   Finally, when we add the additional self-modulated Batch Norm (sBN)~\citep{abn} to SS-GAN,
   which mimics generator conditioning in the unconditional setting,
   this unconditional model attains the same mean performance as the conditional GAN.}
\end{figure*}

\vspace{2mm}
\noindent\textbf{Results}\quad
Figure~\ref{fig:convergence_curves} shows FID training curves on \cifar{} and \imagenet{}.
Table~\ref{tab:best_fid} shows the FID of the best run across three random seeds for each dataset and model combination. The unconditional GAN is unstable on \imagenet{} and the training often diverges. The conditional counterpart outperforms it substantially. The proposed method, namely SS-GAN, is stable on \imagenet{}, and performs substantially better than the unconditional GAN. When equipped with self-modulation it matches the performance on the conditional GAN.
In terms of mean performance (Figure~\ref{fig:convergence_curves}) the proposed approach matches the conditional GAN, and in terms of the best models selected across random seeds (Table~\ref{tab:best_fid}), the performance gap is within $5\%$. On \cifar{} and \lsun{} we observe a substantial improvement over the unconditional GAN and matching the performance of the conditional GAN. Self-supervision appears not to significantly improve the results on \celebahq{}. We posit that this is due to low-diversity in \celebahq{}, and also for which rotation task is less informative.

\vspace{2mm}
\noindent\textbf{Robustness across hyperparameters}\quad
GANs are fragile; changes to the hyperparameter settings have a substantial impact to their performance~\citep{lucic2018,kurach2018gan}.
Therefore, we evaluate different hyperparameter settings to test the stability of SS-GAN.
We consider two classes of hyperparameters:
First, those controlling the Lipschitz constant of the discriminator, a central quantity analyzed in the GAN literature~\citep{ miyato2018spectral,zhou2018understanding}.
We evaluate two state-of-the-art techniques:
gradient penalty~\citep{gulrajani2017improved}, and spectral normalization~\citep{miyato2018spectral}.
The gradient penalty introduces a regularization strength parameter, $\lambda$.
We test two values $\lambda \in \{1,10\}$.
Second, we vary the hyperparameters of the Adam optimizer.
We test two popular settings $(\beta_1, \beta_2)$: $(0.5, 0.999)$ and $(0, 0.9)$.
Previous studies find that multiple discriminator steps per generator step help training~\citep{goodfellow2014generative,salimans2016improved},
so we try both $1$ and $2$ discriminator steps per generator step.

Table~\ref{tab:gp_robustness_fid} compares the mean FID scores of the unconditional models across penalties and optimization hyperparameters. We observe that the proposed approach yields consistent performance improvements.
We observe that in settings where the unconditional GAN collapses (yielding FIDs larger than 100) the self-supervised model does not exhibit such a collapse.

\subsection{Large Scale Self-Supervised GAN}

We scale up training the SS-GAN to attain the best possible FID for unconditional \imagenet{} generation.
To do this, we increase the model's capacity to match the model in~\citep{brock2018large}.\footnote{The details can be found at \url{https://github.com/google/compare\_gan}.}
We train the model on 128 cores of Google TPU v3 Pod for $500$k steps using batch size of 2048. For comparison, we also train the same model without the auxiliary self-supervised loss (Uncond-GAN). We report the FID at $50$k to be comparable other literature reporting results on \imagenet{}. We repeat each run three times with different random seeds.

For SS-GAN we obtain the FID of $23.6 \pm 0.1$ and $71.6 \pm 66.3$ for Uncond-GAN. Self-supervision stabilizes training; the mean and variance across random seeds is greatly reduced because, unlike for the regular unconditional GAN, SS-GAN never collapsed. We observe improvement in the best model across random seeds, and the best SS-GAN attains an FID of $23.4$. To our knowledge, this is the best results attained training unconditionally on \imagenet{}.

\begin{figure}[t]
  \centering
  \includegraphics[width=0.95\textwidth]{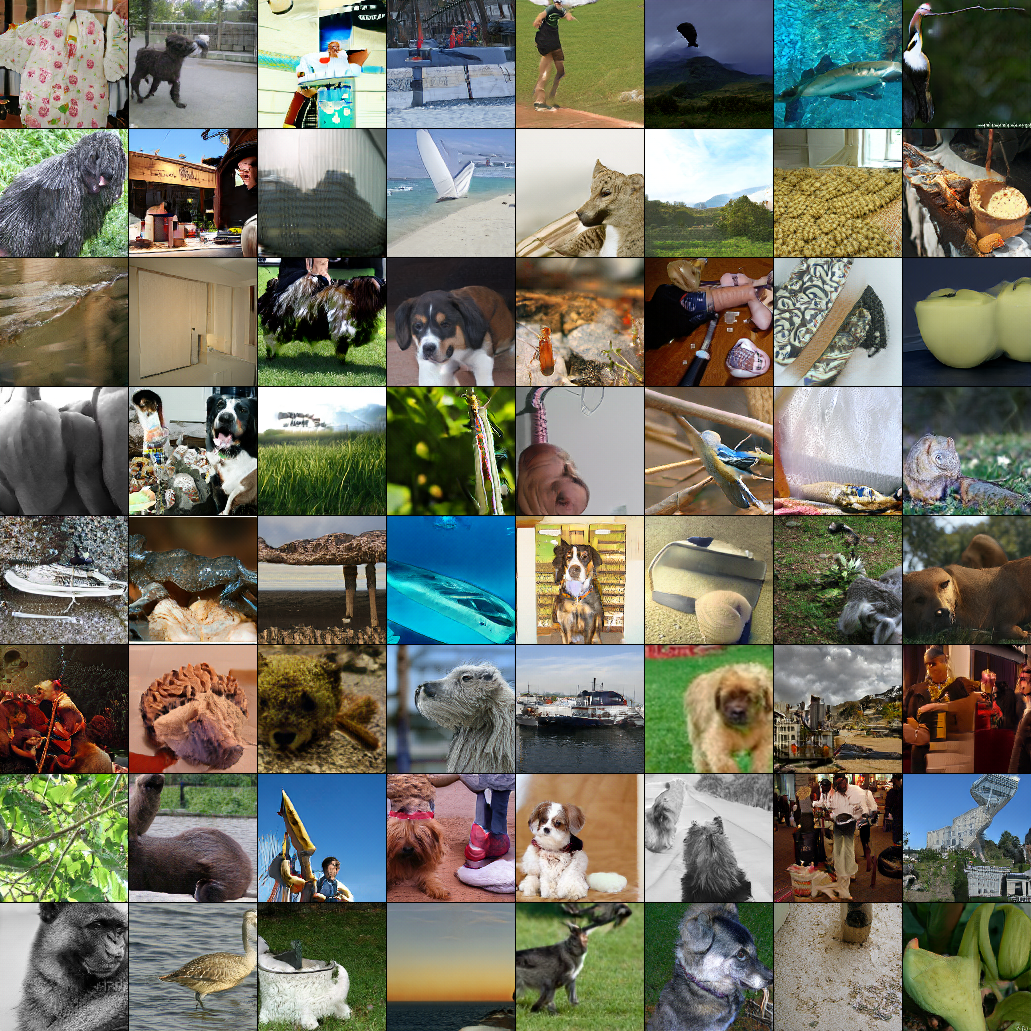}
  \caption{\label{fig:tpu_images}
  A random sample of unconditionally generated images from the self-supervised model.
  To our knowledge, this is the best results attained training unconditionally on \imagenet{}.}
\end{figure}

\subsection{Representation Quality \label{sec:self_sup_eval}}
We test empirically whether self-supervision encourages the discriminator to learn meaningful representations.
For this, we compare the quality of the representations extracted from the intermediate layers of the discriminator's ResNet architecture.
We apply a common evaluation method for representation learning, proposed in~\citet{zhang2016colorful}.
In particular, we train a logistic regression classifier on the feature maps from each ResNet block to
perform the 1000-way classification task on \imagenet{} or 10-way on \cifar{} and report top-1 classification accuracy.

We report results using the Cond-GAN, Uncond-GAN, and SS-GAN models.
We also ablate the adversarial loss from our SS-GAN which results in a purely rotation-based self-supervised model (Rot-only) which uses the same architecture and hyperparameters as the SS-GAN discriminator.
We report the mean accuracy and standard deviation across three independent models with different random seeds.
Training details for the logistic classifier are in the appendix.

\begin{table}[b]
  \centering
  \small
\caption{
\label{tab:cifar10_top1} Top-1 accuracy on \cifar.
Mean score across three training runs of the original model.
All standard deviations are smaller than $0.01$ and are reported in the appendix.
}
\begin{tabular}{ccccc}
  \toprule
       & Uncond. & Cond. & Rot-only & SS-GAN (sBN)           \\
  \midrule
  Block0   & $0.719$    & $0.719$  & $0.710$  & $\mathbf{0.721}$ \\
  Block1   & $0.762$    & $0.759$  & $0.749$  & $\mathbf{0.774}$ \\
  Block2   & $0.778$    & $0.776$  & $0.762$  & $\mathbf{0.796}$ \\
  Block3   & $0.776$    & $0.780$  & $0.752$  & $\mathbf{0.799}$ \\ 
  \midrule
  Best & $0.778$    & $0.780$  & $0.762$  & $\mathbf{0.799}$ \\
  \bottomrule
\end{tabular}

\end{table}

\begin{table}[b]
\centering
\small
\caption{\label{tab:imagenet_top1} Top-1 accuracy on \imagenet{}.
Mean score across three training runs of the original model.
All standard deviations are smaller than $0.01$, except for Uncond-GAN whose results exhibit high variance due to training instability.
All standard deviations are reported in the appendix.
}
\begin{tabular}{ccccc}
\toprule
Method & Uncond.         & Cond.          & Rot-only                    & SS-GAN (sBN)                \\
\midrule
Block0 & $0.074 $           & $0.156$           & $0.147$  & $\mathbf{0.158}$          \\
Block1 & $0.063 $           & $0.187$           & $0.134$           & $\mathbf{0.222}$ \\
Block2 & $0.073 $           & $0.217$           & $0.158$           & $\mathbf{0.250}$  \\
Block3 & $0.083 $           & $0.272$           & $0.202$           & $\mathbf{0.327}$  \\
Block4 & $0.077$           & $0.253$           & $0.196$           & $\mathbf{0.358}$  \\
Block5 & $0.074 $           & $0.337$           & $0.195$           & $\mathbf{0.383}$  \\
\midrule
Best   & $0.083$ & $0.337$ & $0.202$           & $\mathbf{0.383}$  \\
\bottomrule
\end{tabular}

\end{table}

\paragraph{Results}
Table \ref{tab:imagenet_top1} shows the quality of representation at after $1$M training steps on \imagenet{}.
Figure~\ref{fig:accuracy-steps-imagenet} shows the learning curves for representation quality of the final ResNet block on \imagenet{}.
The curves for the other blocks are provided in appendix.
Note that ``training steps'' refers to the training iterations of the original GAN, and not to the linear classifier which is always trained to convergence.
Overall, the SS-GAN yields the best representations across all blocks and training iterations.
We observe similar results on \cifar{} provided  in Table~\ref{tab:cifar10_top1}.

In detail, the \imagenet{} ResNet contains six blocks.
For Uncond-GAN and Rot-only, Block 3 performs best, for Cond-GAN and SS-GAN, the final Block 5 performs best.
The representation quality for Uncond-GAN drops at 500k steps,
which is consistent with the FID drop in Figure~\ref{fig:convergence_curves}.
Overall, the SS-GAN and Cond-GAN representations are better than Uncond-GAN, which correlates with their improved sample quality.
Surprisingly, the the SS-GAN overtakes Cond-GAN after training for $300$k steps.
One possibility is that the Cond-GAN is overfitting the training data.
We inspect the representation performance of Cond-GAN on the training set and indeed see a very large generalization gap, which indicates overfitting.

When we ablate the GAN loss, leaving just the rotation loss, the representation quality substantially decreases.
It seems that the adversarial and rotation losses complement each other both in terms of FID and representation quality.
We emphasize that our discriminator architecture is optimized for image generation, not representation quality.
Rot-only, therefore, is an ablation method,
and is not a state-of-the-art self-supervised learning algorithm.
We discuss these next.

\begin{figure}[t!]
 \centering
 \includegraphics[width=0.9\columnwidth]{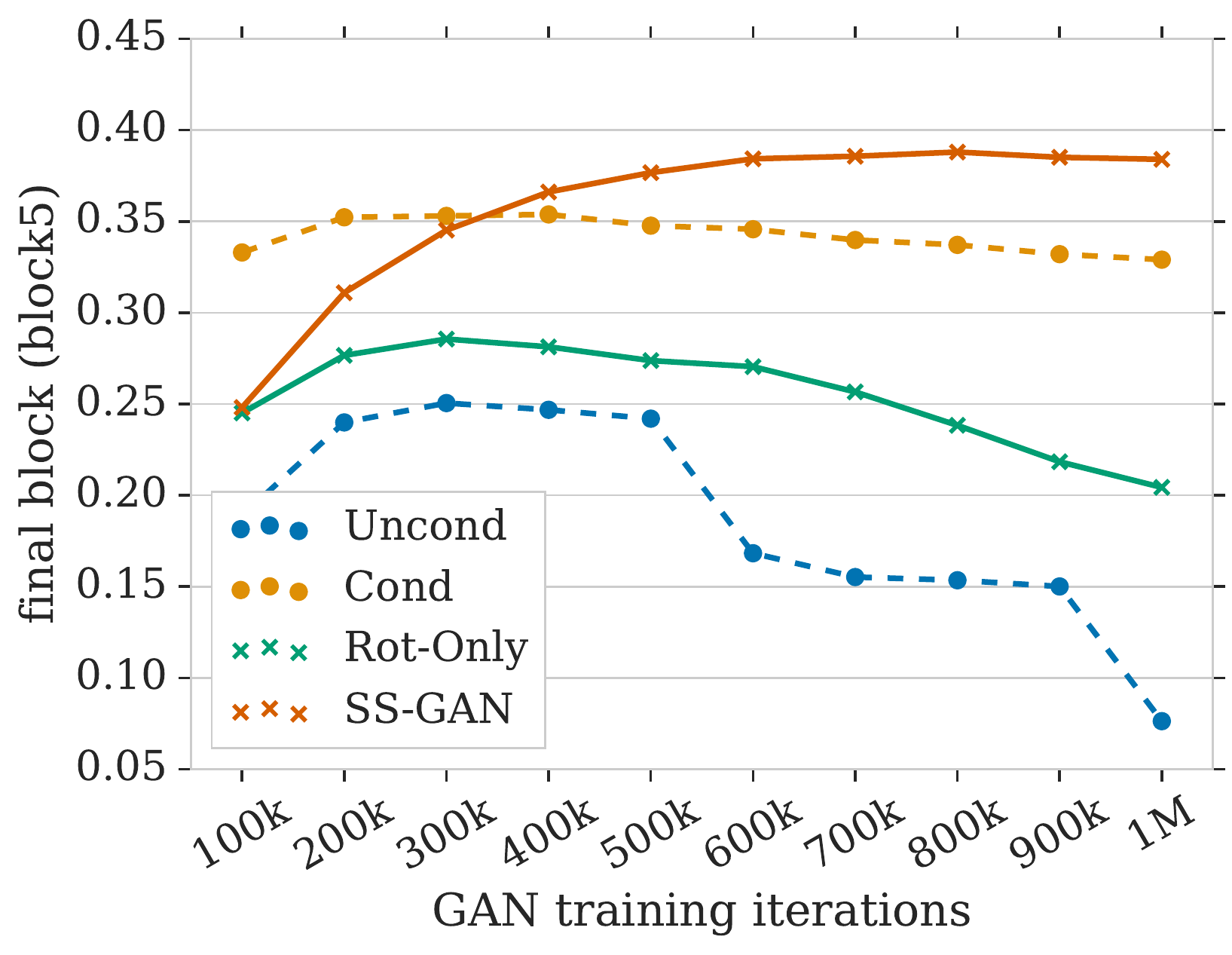}
 \caption{\label{fig:accuracy-steps-imagenet}\imagenet{}
 Top 1 accuracy (mean across three seeds) to predict labels from discriminator representations.
 X-axis gives the number of GAN training iterations.}
\end{figure}

Table~\ref{tab:imagenet_compare} compares the representation quality of SS-GAN to state-of-the-art published self-supervised learning algorithms.
Despite the architecture and hyperparameters being optimized for image quality, the SS-GAN model achieves competitive results on \imagenet{}.
Among those methods, only BiGAN~\citep{donahue2016adversarial} also uses a GAN to learn representations;
but SS-GAN performs substantially (0.073 accuracy points) better.
BiGAN learns the representation with an additional encoder network,
while SS-GAN is arguably simpler because it extracts the representation directly from the discriminator.
The best performing method is the recent DeepClustering algorithm~\citep{caron2018deep}.
This method is just 0.027 accuracy points ahead of SS-GAN and requires expensive offline clustering after every training epoch.

In summary, the representation quality evaluation highlights the correlation between representation quality and image quality.
It also confirms that the SS-GAN does learn relatively powerful image representations.

\begin{table}[h]
\centering
\caption{\label{tab:imagenet_compare} Comparison with other self-supervised representation learning methods by top-1 accuracy on \imagenet{}.
For SS-GAN, the mean performance is presented.}
\begin{tabular}{ll}
  \toprule
  Method                                  & Accuracy \\\midrule
  Context~\citep{doersch2015unsupervised} & 0.317    \\
  BiGAN~\citep{donahue2016adversarial,zhang2017split}    & 0.310    \\
  Colorization~\citep{zhang2016colorful}  & 0.326    \\
  RotNet~\citep{gidaris2018unsupervised}  & 0.387    \\
  DeepClustering~\citep{caron2018deep} & 0.410     \\
  SS-GAN (sBN)                            & 0.383    \\
  \bottomrule
\end{tabular}

\end{table}

\section{Related Work}

\paragraph{GAN forgetting}
Catastrophic forgetting was previously considered as a major cause for GAN training instability.
The main remedy suggested in the literature is to introduce temporal memory into the training algorithm in various ways.
For example, \citet{grnarova2017online} induce discriminator memory by replaying previously generated images.
An alternative is to instead reuse previous models: \citet{salimans2016improved} introduce checkpoint averaging, where a running average of the parameters of each player is kept, and \citet{grnarova2017online} maintain a queue of models that are used at each training iteration.
\citet{kim2018memorization} add memory to retain information about previous samples.
Other papers frame GAN training as a continual learning task.
\citet{thanh2018catastrophic} study catastrophic forgetting in the discriminator and mode collapse, relating these to training instability.
\citet{anonymous2019generative} counter discriminator forgetting by leveraging techniques from continual learning directly (Elastic Weight Sharing~\cite{kirkpatrick2017overcoming} and Intelligent Synapses~\citep{zenke2017continual}).

\vspace{3mm}
\noindent\textbf{Conditional GANs}\quad
Conditional GANs are currently the best approach for generative modeling of complex data sets, such as ImageNet.
The AC-GAN was the first model to introduce an auxiliary classification loss for the discriminator~\citep{odena2017}. The main difference between AC-GAN and the proposed approach is that self-supervised GAN requires no labels. Furthermore, the AC-GAN generator generates images conditioned on the class, whereas our generator is unconditional and the images are subsequently rotated to produce the artificial label. Finally, the self-supervision loss for the discriminator is applied only over real images, whereas the AC-GAN uses both real and fake.

More recently, the P-cGAN model proposed by~\citet{miyato2018cgans} includes one real/fake head per class~\citep{miyato2018cgans}. This architecture improves performance over AC-GAN. The best performing GANs trained on GPUs~\citep{zhang2018self} and TPUs~\citep{brock2018large} use P-cGAN style conditioning in the discriminator. We note that conditional GANs also use labels in the generator, either by concatenating with the latent vector, or via FiLM modulation~\citep{de2017modulating}.

\vspace{3mm}
\noindent\textbf{Self-supervised learning}\quad
Self-supervised learning is a family of methods that learn the high level semantic representation by solving a surrogate task.
It has been widely used in the video domain~\citep{agrawal2015learning,lee2017unsupervised}, the robotics domain~\citep{jang2018grasp2vec,pinto2016supersizing} and the image domain~\citep{doersch2015unsupervised, caron2018deep}. We focused on the image domain in this paper. \citet{gidaris2018unsupervised} proposed to rotate the image and predict the rotation angle. This conceptually simple task yields useful representations for downstream image classification tasks. Apart form trying to predict the rotation, one can also make edits to the given image and ask the network to predict the edited part. For example, the network can be trained to solve the context prediction problem, like the relative location of disjoint patches ~\citep{doersch2015unsupervised, mundhenk2018improvements} or the patch permutation of a jigsaw puzzle~\citep{noroozi2016unsupervised}.
Other surrogate tasks include image inpainting~\citep{pathak2016context}, predicting the color channels from a grayscale image~\citep{zhang2016colorful}, and predicting the unsupervised clustering classes~\citep{caron2018deep}.
Recently, \citet{kolesnikov2019revisiting} conducted a study on self-supervised learning with modern neural architectures.

\section{Conclusions and Future Work}

Motivated by the desire to counter discriminator forgetting, we propose a deep generative model that combines adversarial and self-supervised learning. The resulting novel model, namely self-supervised GAN when combined with the recently introduced self-modulation, can match equivalent conditional GANs on the task of image synthesis, \emph{without having access to labeled data.}
We then show that this model can be scaled to attain an FID of 23.4 on unconditional ImageNet generation which is an extremely challenging task.

This line of work opens several avenues for future research. First, it would be interesting to use a state-of-the-art self-supervised architecture for the discriminator, and optimize for best possible representations. Second, the self-supervised GAN could be used in a semi-supervised setting where a small number of labels could be used to fine-tune the model. Finally, one may exploit several recently introduced techniques, such as self-attention, orthogonal normalization and regularization, and sampling truncation~\citep{zhang2018self,brock2018large}, to yield even better performance in unconditional image synthesis.

We hope that this approach, combining collaborative self-supervision with adversarial training, can pave the way towards high quality, fully unsupervised, generative modeling of complex data.

\subsection*{Acknowledgements}

We would also like to thank Marcin Michalski, Karol Kurach and Anton Raichuk for their help with infustrature, and major contributions to the Compare GAN library.
We appreciate useful discussions with Ilya Tolstikhin, Olivier Bachem, Alexander Kolesnikov, Josip Djolonga, and Tiansheng Yao.
Finally, we are grateful for the support of other members of the Google Brain team, Z\"{u}rich.

{\small
\bibliographystyle{unsrtnat}
\bibliography{gan}
}

\clearpage
\appendix
\section{FID Metric Details}

We compute the FID score using the protocol as described in~\citep{heusel2017gans}.
The image embeddings are extracted from an Inception V1 network provided by the TF library~\citep{tfgan2017},
We use the layer ``pool\_3''.
We fit the multivariate Gaussians used to compute the metric to real samples from the test sets and fake samples.
We use $3000$ samples for \celebahq{} and $10000$ for the other datasets.

\section{SS-GAN Hyper-parameters}

\begin{figure}[h]
 \centering
\begin{subfigure}[b]{0.48\textwidth}
    \includegraphics[width=1\textwidth]{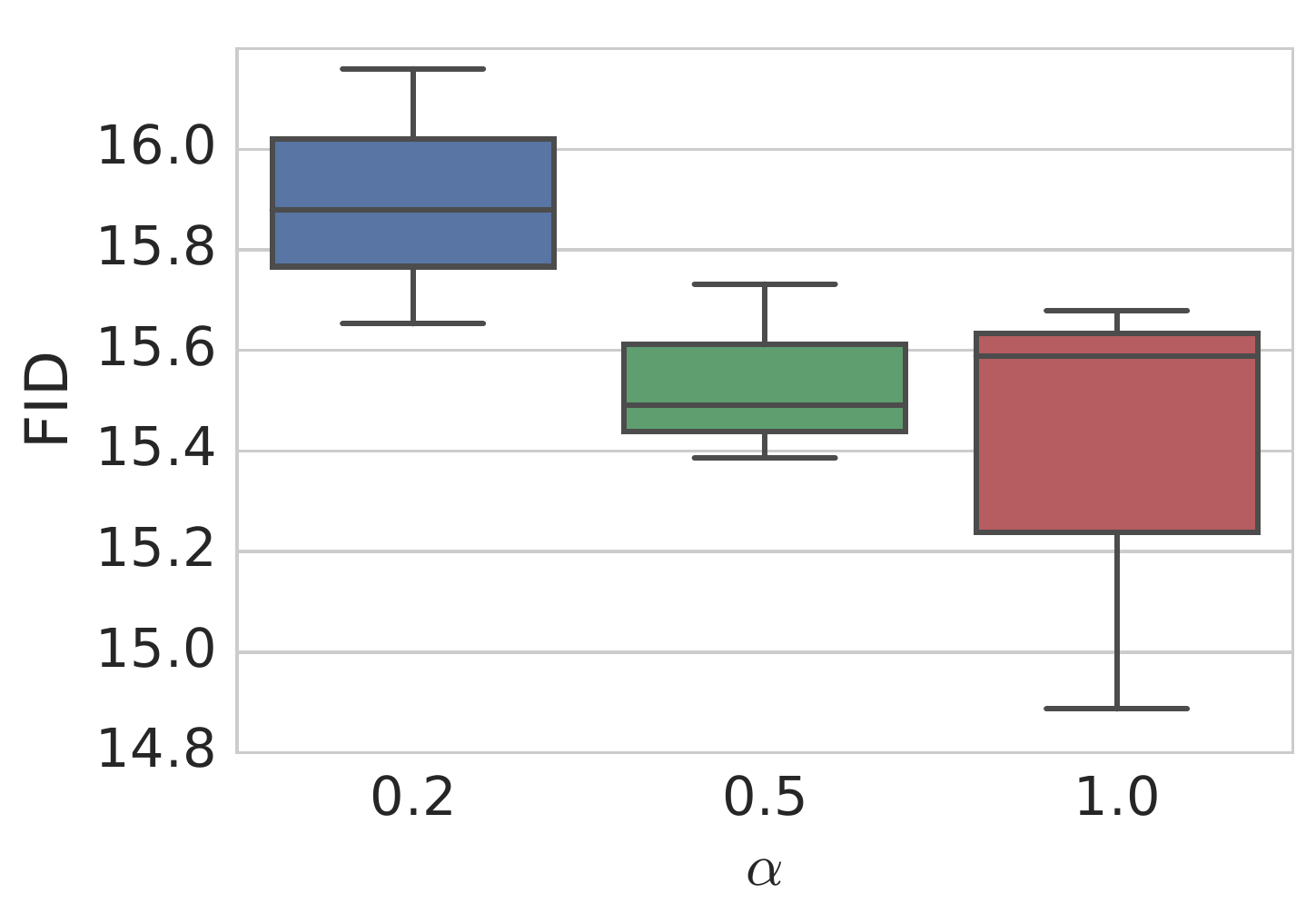}
    \caption{\cifar{}}
 \end{subfigure}
 \begin{subfigure}[b]{0.48\textwidth}
    \includegraphics[width=1\textwidth]{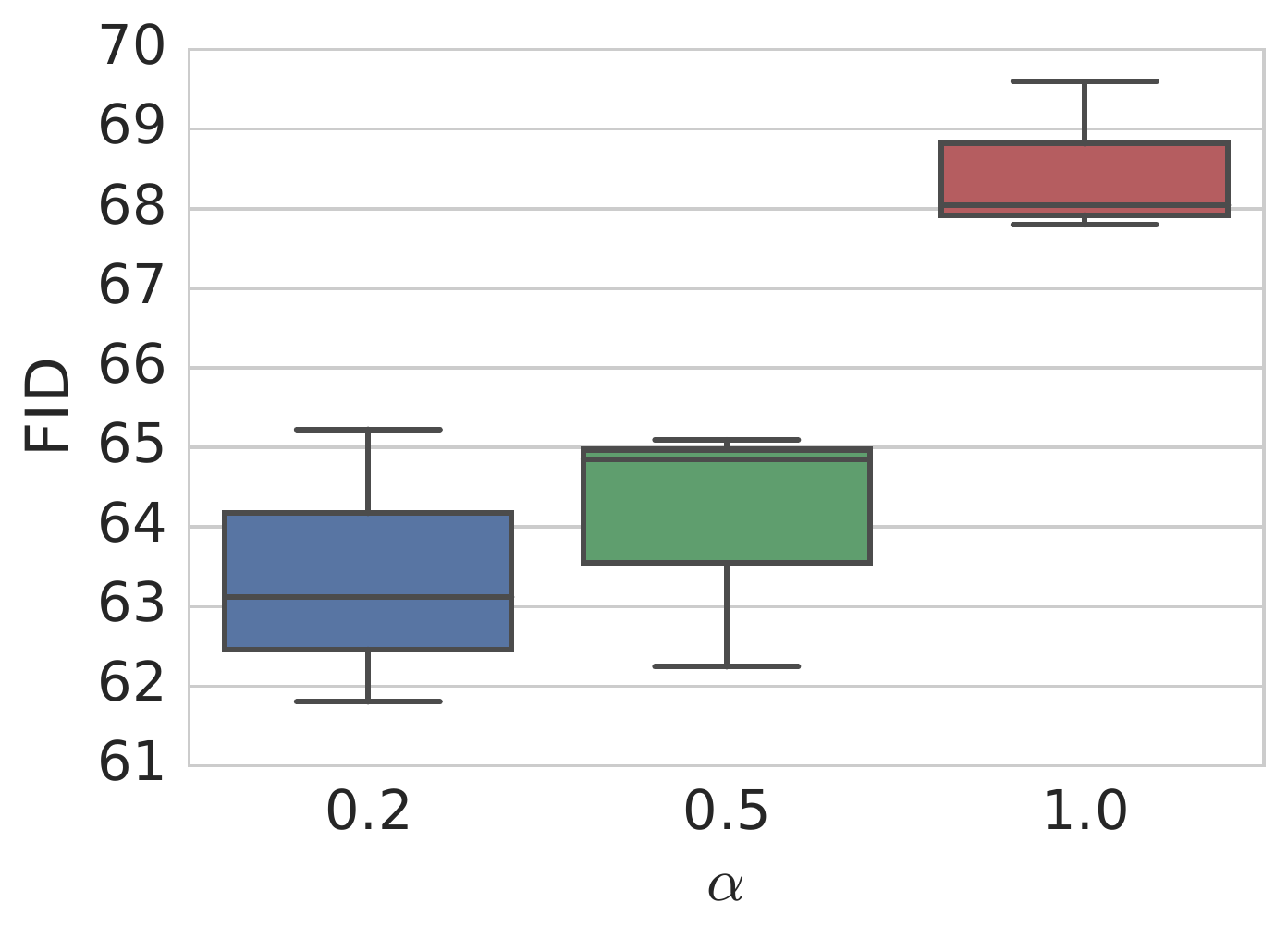}
    \caption{\imagenet{}}
 \end{subfigure}
 \caption{\label{fig:beta_variant} Performance under different $\alpha$ values.}
\end{figure}

We compare different choices of $\alpha$, while fixing $\beta=1$ for simplicity.
A reasonable value of $\alpha$ helps                                                                                                                                              aqa the generator to train using the self-supervision task,
however, an inappropriate value of $\alpha$ could bias the convergence point of the generator.
Table \ref{fig:beta_variant} shows the effectiveness of $\alpha$.
In the values compared, the optimal $\alpha$ is 1 for \cifar{}, and 0.2 for \imagenet{}.
In our main experiments, we used $\alpha=0.2$ for all datasets.

\section{Representation Quality}

\subsection{Implementation Details}
We  train  the  linear  evaluation models with batch size 128 and learning rate of $0.1 \times \frac{\mathrm{batch\_size}}{256}$ following the linear scaling rule~\cite{goyal2017accurate}, for 50 epochs.
The learning rate is decayed by a factor of 10 after epoch 30 and epoch 40.
For data augmentation we resize the smaller dimension of the image to 146 and preserve the aspect ratio.
After that we crop the image to $128 \times 128$.
We apply a random crop for training and a central crop for testing.
The model is trained on a single NVIDIA Tesla P100 GPU.

\subsection{Additional Results}
Table~\ref{tab:cifar10_top1_full} shows the top-1 accuracy with on \cifar{} with standard deviations.
The results are stable on \cifar{} as all the standard deviation is within 0.01.
Table~\ref{tab:imagenet_top1_full} shows the top-1 accuracy with on \imagenet{} with standard deviations.
Uncond-GAN representation quality shows large variance as we observe that the unconditional GAN collapses in some cases.

\begin{figure*}[h]
\begin{center}
\epsfig{file=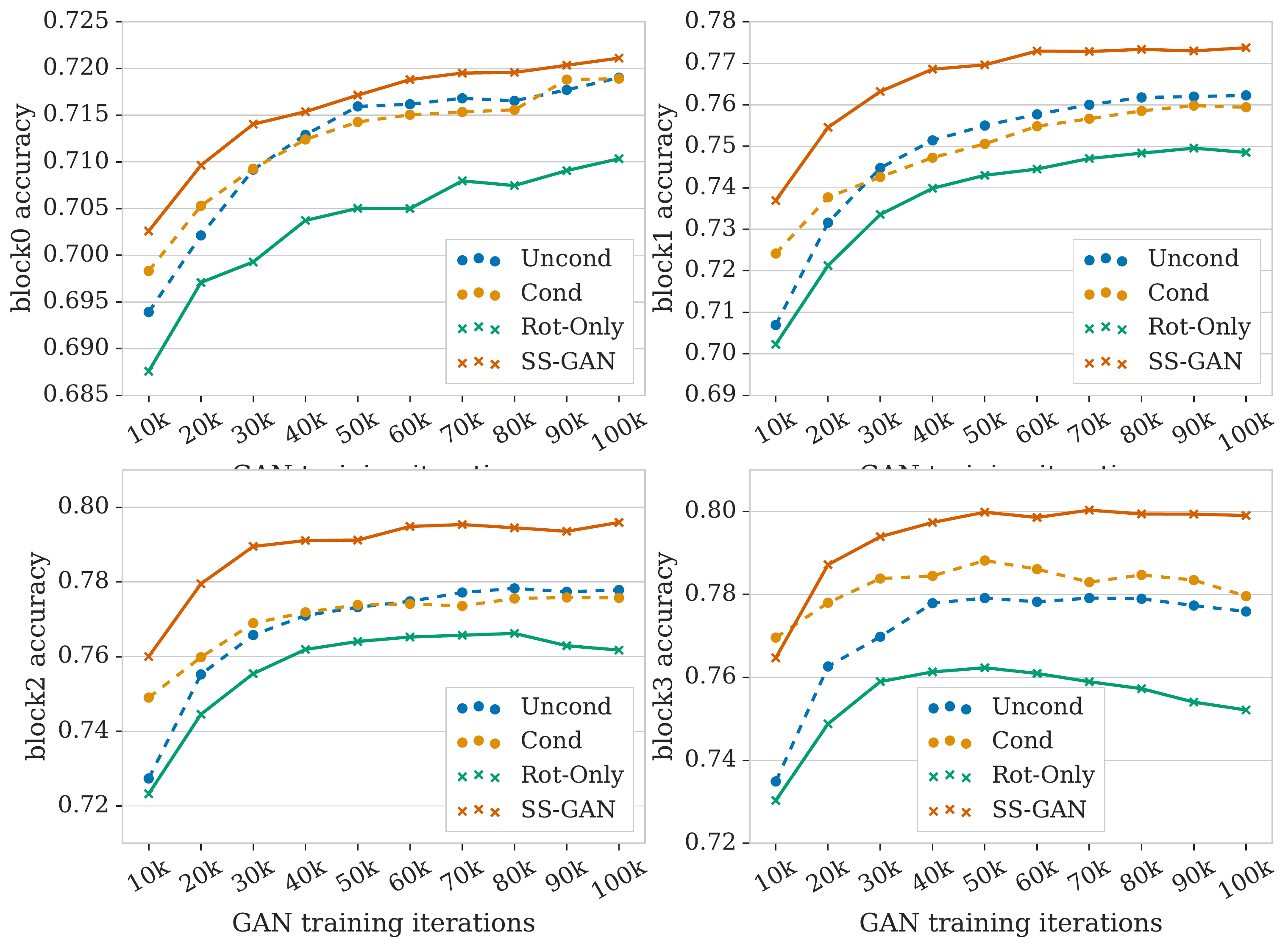,height=12.0cm}
\end{center}
\caption{\label{fig:accuracy-steps-cifar} Top 1 accuracy on \cifar{} with training steps from 10k to 100k.}
\end{figure*}

Figure~\ref{fig:accuracy-steps-cifar} shows the representation quality on all 4 blocks on the \cifar{} dataset.
SS-GAN consistently outperforms other models on all 4 blocks.
Figure~\ref{fig:accuracy-steps-imagenet} shows the representation quality on all 6 blocks on the \imagenet{} dataset.
We observe that all methods perform similarly before 500k steps on block0, which contains low level features.
While going from block0 to block6, the conditional GAN and SS-GAN achieve much better representation results.
The conditional GAN benefits from the supervised labels in layers closer to the classification head.
However, the unconditional GAN attains worse result at the last layer and the rotation only model gets decreasing quality with more training steps.
When combining the self-supervised loss and the adversarial loss, SS-GAN representation quality becomes stable and outperforms the other models.

\begin{figure*}[h]
\begin{center}
\epsfig{file=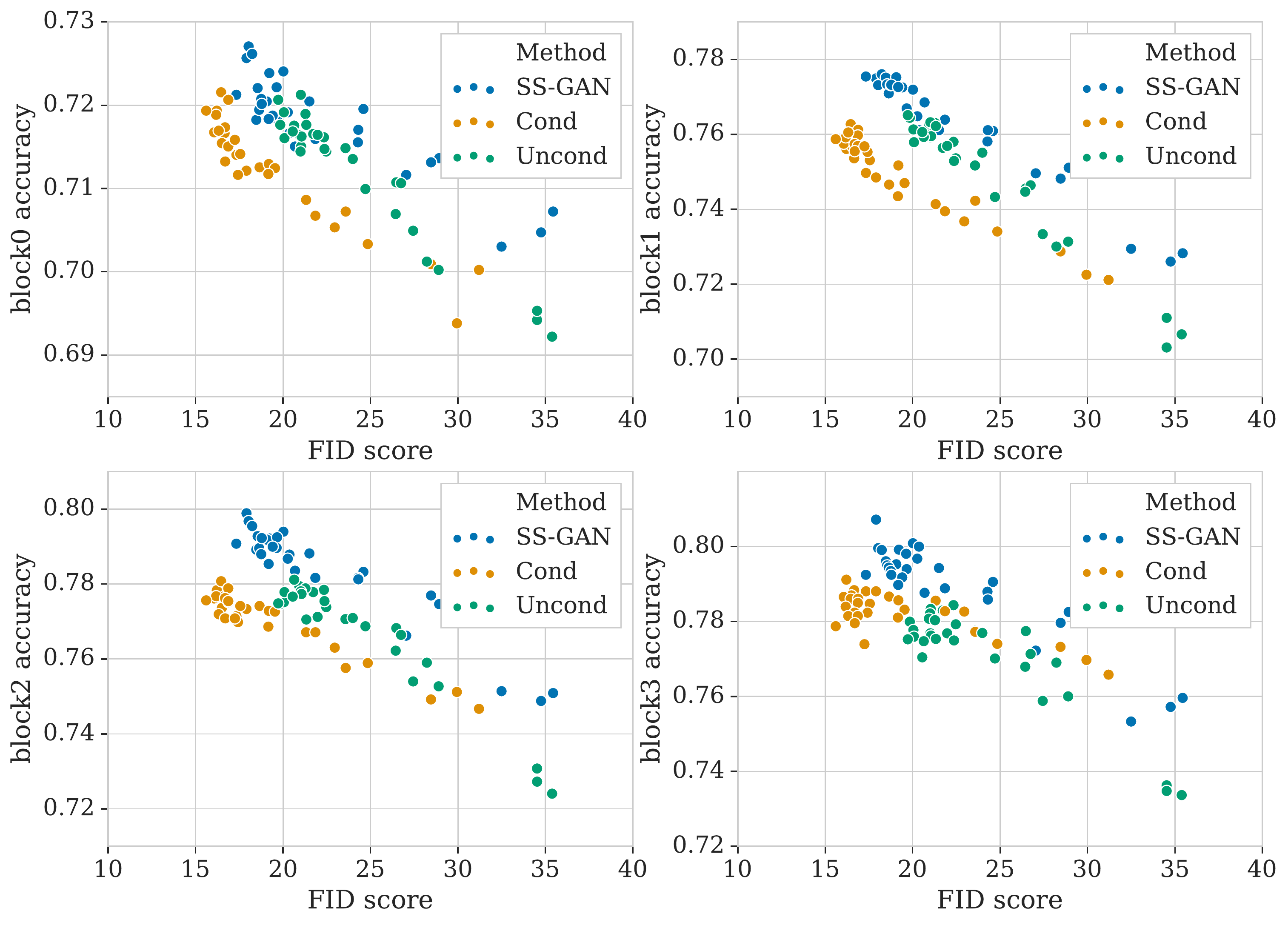,height=12.0cm}
\end{center}
\caption{\label{fig:accuracy-fid-cifar}
Correlation between top-1 accuracy and FID score for different numbers of GAN training steps from 10k to 100k on \cifar{}.
Overall, the representation quality and the FID score is correlated for all methods.
The representation quality varies up to 4\% with the same FID score.
}
\end{figure*}

\begin{figure*}[h]
\begin{center}
\epsfig{file=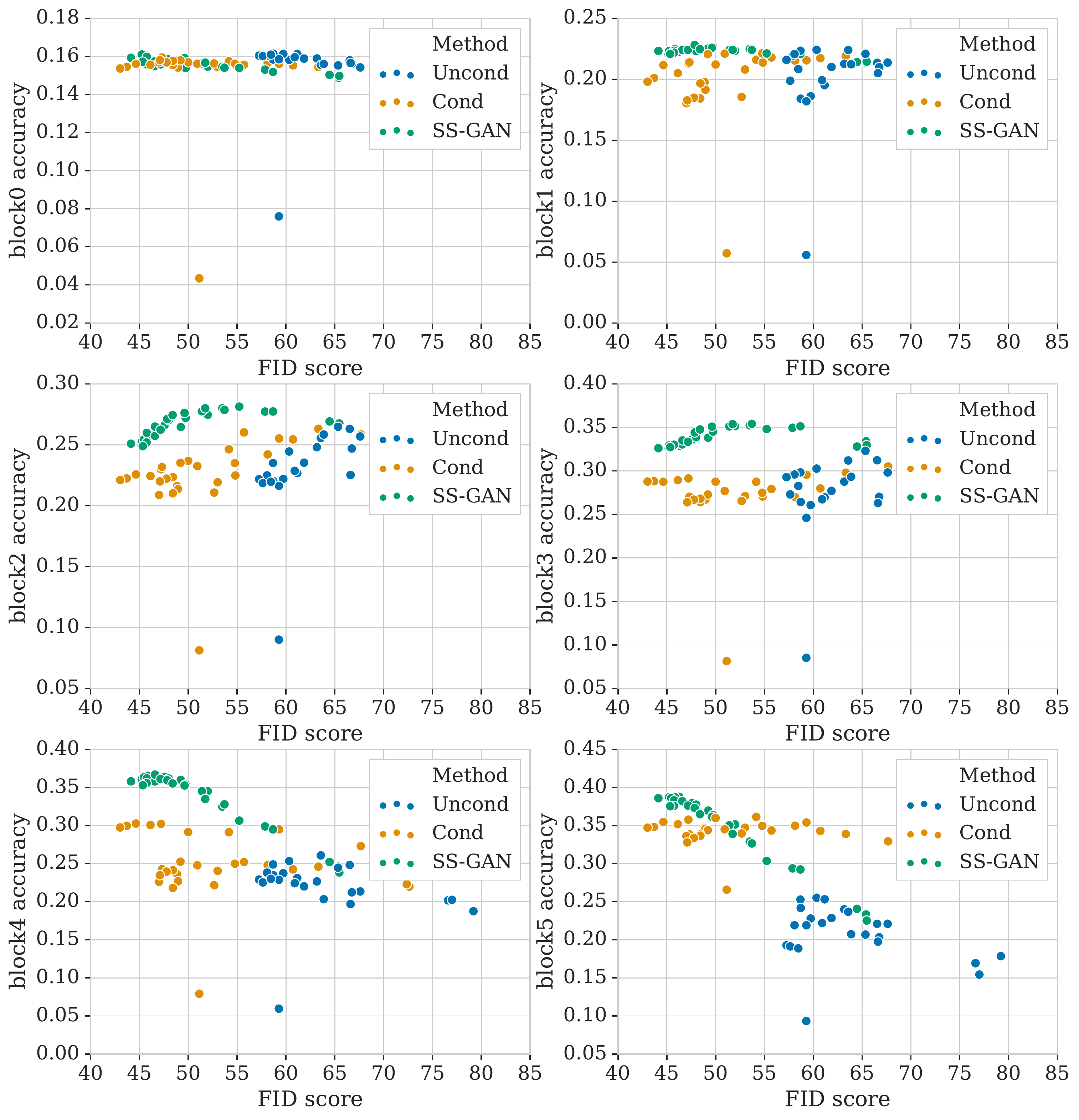,height=18.0cm}
\end{center}
\caption{\label{fig:accuracy-fid-imagenet}
Correlation between top-1 accuracy and FID score for different numbers of GAN training steps from 100k to 1M on \imagenet{}.
Representation quality and FID score are not correlated on any of block0 to block4.
This indicates that low level features are being extracted, which perform similarly on the \imagenet{} dataset.
Starting from block4, SS-GAN attains better representation as the FID score improves.
}
\end{figure*}

Figure~\ref{fig:accuracy-fid-cifar} and Figure~\ref{fig:accuracy-fid-imagenet} show the correlation between top-1 accuracy and FID score.
We report the FID and top-1 accuracy from training steps 10k to 100k on \cifar{}, and 100k to 1M on \imagenet{}.
We evaluate $10 \times 3$ models in total, where 10 is the number of training steps at which we evaluate and 3 is the number of random seeds for each run.
The collapsed models with FID score larger than 100 are removed from the plot.
Overall, the representation quality and the FID score is correlated for all methods on the \cifar{} dataset.
On \imagenet{}, only SS-GAN gets better representation quality with better sample quality on block4 and block5.

\begin{table*}
  \centering
  \caption{\label{tab:cifar10_top1_full} Top-1 accuracy on \cifar{} with standard variations.}
  \begin{tabular}{llllll}
\toprule
Method & Uncond-GAN & Cond-GAN & Rot-only & SS-GAN (sBN) \\
\midrule
Block0    &   $0.719 \pm 0.002$ &   $0.719 \pm 0.003$ & $0.710 \pm 0.002$ & $\mathbf{0.721 \pm 0.002}$\\
Block1      &   $0.762 \pm  0.001$ &   $0.759 \pm  0.003$ & $0.749 \pm 0.003$ & $\mathbf{0.774 \pm 0.003}$ \\
Block2 &   $0.778 \pm  0.001$ &   $0.776 \pm  0.005$ & $0.762 \pm 0.003$ & $\mathbf{0.796 \pm 0.005}$ \\
Block3  &  $0.776 \pm  0.005$ &   $0.780 \pm  0.006$ & $0.752 \pm 0.006$ & $\mathbf{0.799 \pm 0.003}$ \\
\midrule
Best & $0.778 \pm  0.001$ & $0.780 \pm  0.006$ & $0.762 \pm 0.003$ & $\mathbf{0.799 \pm 0.003}$ \\
\bottomrule
\end{tabular}

\end{table*}

\begin{table*}
  \centering
  \caption{\label{tab:imagenet_top1_full} Top-1 accuracy on \imagenet{} with standard variations.}
  \begin{tabular}{lllll}
\toprule
Method & Uncond-GAN & Cond-GAN & Rot-only & SS-GAN (sBN) \\
\midrule
Block0      &    $0.074 \pm 0.074$   &    $0.156 \pm  0.002$    &    $0.147 \pm  0.001$    &   $\mathbf{0.158 \pm  0.001}$ \\
Block1      &    $0.063 \pm  0.103$   &    $0.187 \pm  0.010$    &    $0.134 \pm  0.003$    &   $\mathbf{0.222 \pm  0.001}$ \\
Block2      &    $0.073 \pm  0.124$   &    $0.217 \pm 0.007 $    &    $0.158 \pm 0.003$     &   $\mathbf{0.250 \pm 0.001}$ \\
Block3      &    $0.083 \pm 0.142 $   &    $0.272 \pm 0.014$    &    $0.202 \pm 0.005$     &   $\mathbf{0.327 \pm 0.001}$ \\
Block4      &    $0.077 \pm 0.132$   &    $0.253 \pm 0.040 $    &    $0.196 \pm 0.001$     &   $\mathbf{0.358 \pm 0.005}$ \\
Block5      &    $0.074 \pm 0.126$   &    $0.337 \pm 0.010$    &    $0.195 \pm 0.029$     &   $\mathbf{0.383 \pm 0.007}$ \\
\midrule
Best        &    $0.083 \pm 0.142 $     & $0.337 \pm 0.010$ & $0.202 \pm 0.005$ & $\mathbf{0.383 \pm 0.007}$ \\
\bottomrule
\end{tabular}

\end{table*}

\begin{figure*}[h]
\begin{center}
\epsfig{file=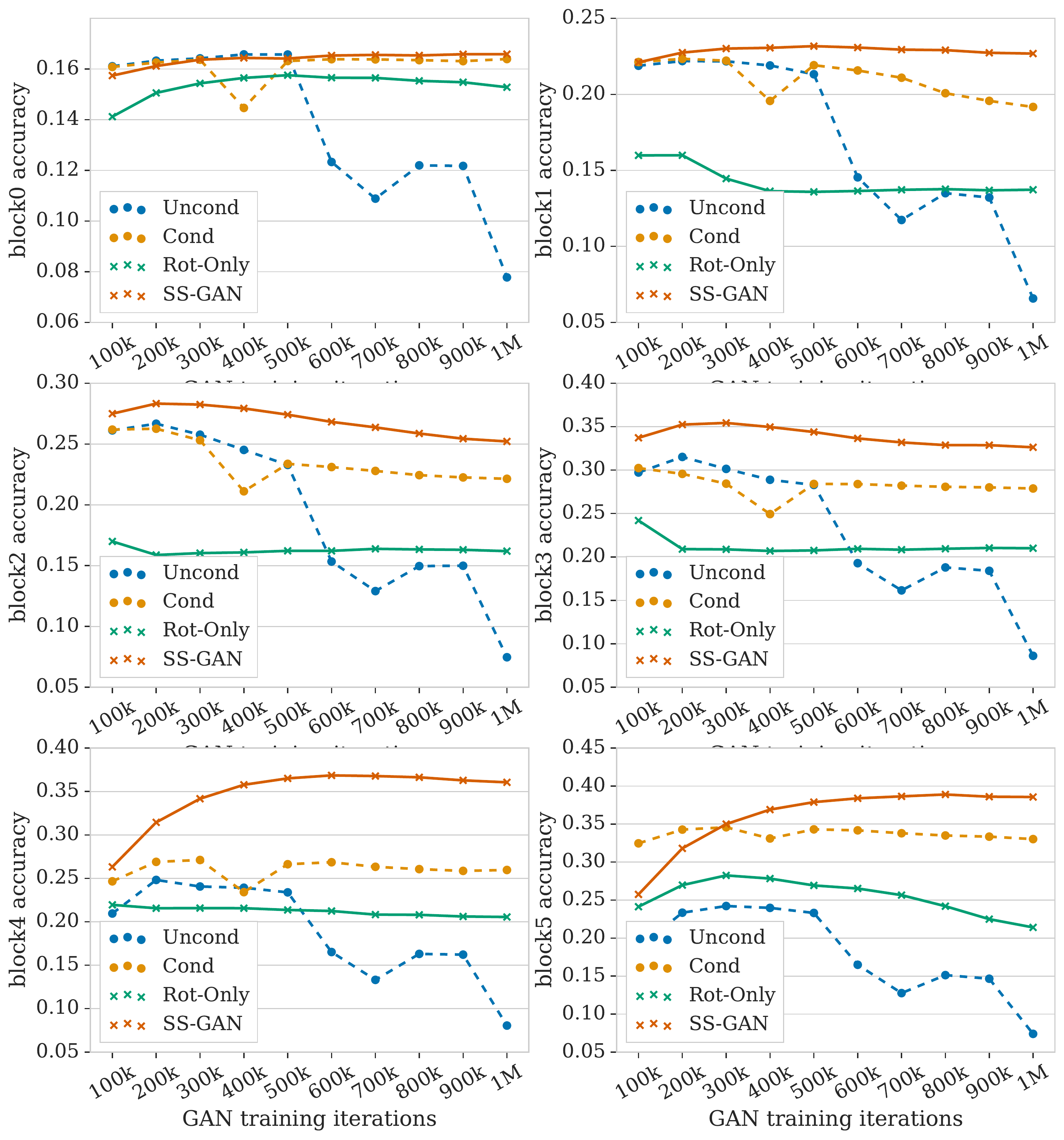,height=18.0cm}
\end{center}
\caption{\label{fig:accuracy-steps-imagenet} Top 1 accuracy on \imagenet{} validation set with training steps from 10k to 1M.
}
\end{figure*}

\end{document}